\documentclass{article}
\usepackage{spconf,amsmath,graphicx}
\usepackage{algorithm2e}
\usepackage{listings}
\usepackage{verbatim}
\usepackage{amsmath}
\usepackage{graphicx}
\usepackage{amssymb}
\usepackage{qcircuit}
\usepackage{mathtools}
\usepackage{tikz}
\usepackage{pgfplots}
\pgfplotsset{
  compat=newest,
  xlabel near ticks,
  ylabel near ticks
}

\DeclarePairedDelimiter\ket{\lvert}{\rangle}
\DeclarePairedDelimiterX\braket[2]{\langle}{\rangle}{#1 \delimsize\vert #2}
\title{IMAGE PROCESSING IN QUANTUM COMPUTERS}
\name{Aditya Dendukuri and Khoa Luu}
\address{\textit{adenduku@email.uark.edu, khoaluu@uark.edu}\\
         $^{1}$Computer Vision and Image Understanding Lab \\Computer Science and Computer Engineering Department, \\
         University of Arkansas, Fayetteville, AR, 72701\\}
         
\begin{document}
%\ninept
%
\maketitle
\begin{abstract}
\textit{Quantum Image Processing (QIP)} is an exciting new field showing a lot of promise as a powerful addition to the arsenal of Image Processing techniques. Representing image pixel by pixel using classical information requires enormous amount of computational resources. Hence, exploring methods to represent images in a different paradigm of information is important. In this work, we study the representation of images in \textit{Quantum Information}. The main motivation for this pursuit is the ability of storing N bits of classical information in only $\log_2{N}$ quantum bits (qubits). The promising first step was the exponentially efficient implementation of the Fourier transform in quantum computers as compared to Fast Fourier Transform in classical computers.
In addition, Images encoded in quantum information could obey unique quantum properties like superposition or entanglement. 

\end{abstract}
\begin{keywords}
Quantum Image Processing, Quantum Computer Vision, Quantum Fourier Transform
\end{keywords}

\section{Introduction}
\textit{Image Processing} is a very well established field in Computer Science with many applications in the modern world such as facial recognition \cite{16, 17, 18}, image analysis \cite{19, 20, 23, 24}, image segmentation \cite{21, 22} and de-noising using a large arsenal of techniques. However, current image processing methods demand expensive computational resources to store and process images. There are many computational techniques like the Fast Fourier Transform (FFT) that provide a decent speedup for image processing. \textit{Quantum Computing} on the other hand, defines a probabilistic approach to represent classical information using methods from Quantum Theory. The idea of quantum information was first introduced in 1980 by Paul Benioff \cite{1}. In the same year, Yuri Manin proposed a quantum computer in his textbook ``Computable and Uncomputable" \cite{2}. In 1982, the field was formalized and made popular by Richard Feynman in his paper about simulating physics in computers \cite{3}. David Deutch further advanced the field by formulating a quantum turing machine \cite{4}. Thanks to these advancements, quantum computing showed a lot of promise for the future of computing. This new class of decision problems are also grouped in a computational class called \textit{bounded-error quantum polynomial time (BQP)}. The core idea of quantum computing is a \textit{qubit}: the quantum analogue of a classical computer bit. A classical bit is only capable of storing a determined value (0 or 1). A qubit can store both 0 and 1 in an \textit{uncertain} state as a superposition:
\begin{equation}
    \alpha\ket{0} + \beta\ket{1}
\end{equation}
$\alpha$ and $\beta$ are the normalized \textit{probability amplitudes}. 

If we consider N classical bits, we can use the superposition principle to encode classical N bit information in $\log_2N$ quantum bits. However, this advantage comes with its own challenges. For starters, there are very few real quantum computers in existence providing only a handful of qubits as manufacturing qubits require very sensitive materials that have to be stored in extreme environments. Quantum behavior can be simulated on a classical computer but the computational cost increases exponentially with the number of qubits. \textit{Quantum Image Processing (QIP)} is a promising new research topic seeking to exploit the advantages of quantum computers discussed in more detail in section 1.1. There have already been a handful of quantum image models proposed along with quantum analogues of image processing techniques like the \textit{Quantum Fourier Transform (QFT)} \cite{5} and the \textit{Quantum Wavelet Transform (QWT)} \cite{6} . These techniques can be implemented very efficiently in a quantum circuit, hence making them much more efficient than their classical counterparts. Quantum Computers have also shown extraordinary performance in Shor's factoring algorithm \cite{7} and the Grover's search algorithm \cite{8}. There are also machine learning techniques like gradient descent ported to a quantum computer framework. Therefore, these techniques imply a bright future for computationally expensive fields like Image Processing and Computer Vision. In the following sections, we survey a number of quantum image models and demonstrate one of the algorithms by implementing it in a classical computer. 

\section{Why Quantum Computers?}
In this section, we lay out the case for employing quantum information by encoding $2^3$ bits of binary information in $3$ \textit{qubits} (denoted by $\psi$). The key feature of a qubit is the ability to store an \textit{uncertain} value as a superposition of ``0'' or ``1''. In other words, a quantum bit encodes information in such a way that it holds either 0 or 1. Therefore, any operation will intrinsically perform on both the possibilities of the bit being 0 or 1 as shown in Table \ref{table:1}.
\begin{table} [h!]
\centering
 \begin{tabular}{|c|c|c|c|c|c|c|c|c|} 
 \hline
 Qubits & $\phi_1$ & $\phi_2$ & $\phi_3$ & $\phi_4$ & $\phi_5$ & $\phi_6$ & $\phi_7$ & $\phi_8$\\ 
 \hline
 $\psi_1$ & 0 & 0 & 0 & 0 & 1 & 1 & 1 & 1\\ 
 \hline
 $\psi_2$ & 0 & 0 & 1 & 1 & 0 & 0 & 1 & 1\\ 
 \hline
 $\psi_3$ & 0 & 1 & 0 & 1 & 0 & 1 & 0 & 1\\ 
 \hline
\end{tabular}
\caption{All possible states ($\phi_1 , \phi_2$ ...) of a 4 qubit register}
\label{table:1}
\end{table}

The three bit \textit{quantum register} ($\Psi = \psi_1 \psi_2 \psi_3 $) can store $2^3 = 8$ bits of information in a superposition where every configuration ($\phi_i$) is assigned a \textit{probability amplitude} as shown below:
\begin{equation}
    \Psi =  \alpha_0 \ket{000} + \alpha_1 \ket{001} + \alpha_2 \ket{010} ....... \alpha_{2^n-1} \ket{111}
    \label{eq1}
\end{equation}
The probability amplitudes ($\alpha_1,\alpha_2...$) represent the probability of the quantum register to be in that configuration.  It is also very important to note that the quantum logic gates have some unique luxuries that classical systems cannot enjoy. For example, the 2 qubit CNOT gate as shown below:

\[ 
    CNOT = \begin{bmatrix}
    1 & 0 & 0 & 0 \\
    0 & 1 & 0 & 0 \\
    0 & 0 & 0 & 1 \\
    0 & 0 & 1 & 0 \\
    \end{bmatrix}
\]
is self reversible. This means if this gate is applied twice, the qubit will return to the original state. For FRQI, we can exploit this important gate to build the circuit for preparing the quantum image. The CNOT gate can be used as a conditional gate as it only flips the target bit if the input is 1. We can also attach any other unitary operation to the conditional gate instead of the NOT gate to implement a conditional operation. For example, the conditional phase shift gate would look like: 
\[ 
    \begin{bmatrix}
    1 & 0 & 0 & 0 \\
    0 & 1 & 0 & 0 \\
    0 & 0 & 1 & 0 \\
    0 & 0 & 0 & e^{i\phi} \\
    \end{bmatrix}
\]
The three qubit extension of the CNOT gate (also called the Toffoli gate) is \textit{universal}. This means we can reduce any operation possible on a quantum computer to a sequence of these universal gates. 
\section{Quantum Image Models}
Venegas-Andraca and Bose \cite{9} introduced image representation on the quantum computers by proposing the ‘qubit lattice’ method, where each pixel was represented in its quantum state and then a quantum matrix was created with them. However this is a mere quantum analogue of a classical image and there is no added advantage in the quantum form. However, the next huge advancement is work by Le et al.~who provided a flexible representation of quantum images (FRQI) for multiple intensity levels \cite{10}. The FRQI representation is expressed mathematically as follows: 
\begin{equation}
    \ket{I(\theta)}=\frac{1}{2^n}\sum\limits_{i=0}^{2^{2n}-1} (\sin(\theta_i)\ket{0}+\cos(\theta_i)\ket{1}) \ket{i}\\
\end{equation}

Where $\theta_i$ corresponds to the intensity of the $i^{th}$ pixel. Since the intensity values are encoded in the $amplitudes$ of the quantum state, it's relatively straightforward to apply various transformation like QFT as they are applied directly on the amplitudes of the image. Zhang et al. \cite{11} provided a different approach by representing the image pixel values in the $basis$ states instead of the amplitudes called \textit{novel enhanced quantum representation (NEQR)} as shown below:

\begin{equation}
    | I \rangle = \frac { 1 } { 2 ^ { n } } \sum _ { X = 0 } ^ { 2 ^ { n } - 1 } \sum _ { Y = 0 } ^ { 2 ^ { n } - 1 } | f ( X , Y ) \rangle | X Y \rangle
\end{equation}

Where $f ( X , Y )$ refers to the pixel intensity at (X, Y). FRQI and NEQR are pretty comprehensive but they have their own disadvantages. The first main disadvantage is that they require square images ($2^N \times 2^N$) along with needing extra qubits to encode the positions along with the pixel intensities. In addition, it is very tricky to retrieve the classical version of the image from the quantum image due to the added uncertainty. Indeed, images are retrieved via normalized probability distributions. Srivastava et.al. propose a Quantum Image Representation Through Two-Dimensional Quantum States and Normalized Amplitude (2D-QSNA) [12]. This approach can be used to represent rectangular images and the intensity of a pixel without using additional qubits. Therefore, an image with dimensions ($2^N \times 2^M$) can be encoded (position and intensity) in only M + N qubits. The various discussed models can be seen in the comparison in the Table  \ref{Tab1:Compare}. 

\begin{table}[ht]
\centering % used for centering table
\begin{tabular}{|c|c c c|} % centered columns (4 columns)
\hline %inserts double horizontal lines
\textbf{Cases} & \textbf{FRQI} & \textbf{NEQR} & \textbf{2D-QSNA} \\ [0.5ex] % inserts table
%heading
\hline % inserts single horizontal line
Shape & $2^N \times 2^N$ & $2^N \times 2^N$ & $2^N \times 2^M$ \\ % inserting body of the table
Num Qubits & 2N+1 & 2N+l & M+N \\
Complexity & O($2^{4m}$) & O($2^{m}$) & pure state\\
[1ex] % [1ex] adds vertical space
\hline %inserts single line
\end{tabular}
\caption{Comparison of Quantum Image Models}
\label{Tab1:Compare} % is used to refer this table in the text
\end{table}

\subsection{Preperation of FRQI}
FQRI can be prepared by Hadamard and Controlled rotation operators \cite{10}. For an $2^n \times 2^n$, the first step is to initialize the n qubits as $\ket{0}^{\otimes2n+1}$. Next, we apply the Hadamard gate to each qubit to put them in superposition. 

\begin{equation}
    \mathcal { H } ( \ket{0} ^ { \otimes 2 n + 1 } ) = \frac { 1 } { 2 ^ { n } } \ket{0} \otimes \sum _ { i = 0 } ^ { 2 ^ { 2 n } - 1 } \ket{i} 
\end{equation}

Next we apply the controlled rotation operators to each basis state resulting in the joint state as follows: 

\begin{equation}
    \mathcal { R } | H \rangle = \left( \prod _ { i = 0 } ^ { 2 ^ { 2 n } - 1 } R _ { i } \right) | H \rangle = | I ( \theta ) \rangle
\end{equation}

Each of the $R_i$ rotation corresponds to the respective pixel's rotation. Hence we encode the intensities via rotating these quantum states. The circuit implementation can be achieved by combining the Hadamard, phase shift gate, and the  CNOT gate.

\subsection{Quantum Fourier Transform (QFT)}
The Quantum Fourier Transform is the Quantum analogue of the Discrete Fourier Transform (DFT). For QFT, the DFT is applied to the \textit{probability amplitudes} of the quantum state. Mathematically, it is defined as the following transformation: 

\begin{equation}
    \sum_{j}\alpha_j\ket{j}  = \sum_{k}\bar{\alpha_k}\ket{k}
\end{equation}

\begin{equation}
    \bar{\alpha_k} = \frac{1}{\sqrt{N}}\sum_{j=0}^{N-1}e^{({{\frac{2\pi ijk}{N}}})}{\alpha_j}
\end{equation}

The QFT can be implemented in the circuit level by rearranging the terms into a product form which is easy to implement in a circuit \cite{13}. The circuit has proven to be very efficient by its application in Shor's algorithm \cite{7}. The circuit is shown in fig \ref{qftcircuit}. The $R_n$ gate, is a controlled rotation operator acting on the control qubit as shown below:

\begin{equation}
    R _ { n } = \left[ \begin{array} { c c } { 1 } & { 0 } \\ { 0 } & { \exp \left( \frac { - 2 \pi i } { 2 ^ { n } } \right) } \end{array} \right]
\end{equation}

\begin{figure}
    \centering
    \includegraphics[scale = 0.35]{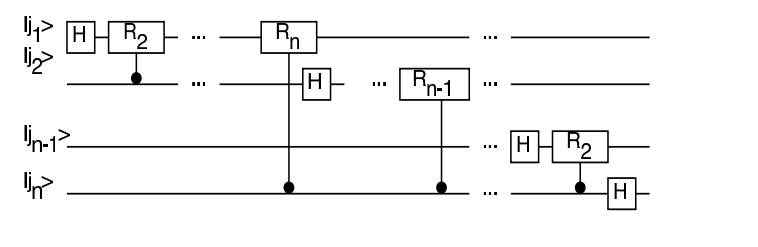}
    \caption{QFT circuit}
    \label{qftcircuit}
\end{figure}

\section{Implementation and Results}
We chose to implement FRQI due to it's simple and effective feature that we are storing the image intensities as the probability amplitudes of the quantum state. Accessing these amplitudes would be associated with measurement but how do we get to this state in the first place? It is proved in \cite{10} that we can achieve the FRQI state using the Hadamard and controlled rotation gates. Interestingly enough, the same type of gates are also used in the QFT circuit. We implemented this both in a classical computer and  on the IBM Quantum Experience cloud. Simulating quantum systems in a classical computer is pretty expensive computationally. This is because we need to explicitly define the superposition (in other words store the superposition in memory). In a real quantum computer, the superposition is intrinsically a property of the system hence no requirement of storing it. For example, consider we would like to encode N bits of information in $\log_2{N}$ bits, the superposition of all the possible combinations of the 0s and 1s can be stored in the quantum register unlike a quantum register in which we can only store one of all the possible states. A classical computer would have to store all the possible states as a list (taking up extra space) unlike a quantum computer. This also means that we do not get the luxury of accessing these quantum bits as in a classical system. The n qubits are the only entities that we can interact with using quantum gates and measurement. The encoded $2^n$ bits of information within these n qubits can only be accessed via measurement. The state we would like to reach is equation 3. To build the circuit we use a 2 x 2 image as shown below:
\[
\begin{bmatrix}
\theta_1 & \theta_2 \\
\theta_3 & \theta_4 \\
\end{bmatrix}
\]
We will encode the information as:\\ \\ 
$\frac{1}{2}[(cos(\theta_1)+sin(\theta_1))\ket{00}+(cos(\theta_2)+sin(\theta_2))\ket{01}+(cos(\theta_3)+sin(\theta_3))\ket{10}+(cos(\theta_4)+sin(\theta_4))\ket{11}]$\\

This encoding is possible by using  the controlled rotation operator shown in figure 2, to rotate the individual basis states based on the intensities. We tested this algorithm on a 2x2 image, The IBM quantum experience machine does handle enough qubits to hold an image larger than 2x2. Therefore, we present the mechanics of the circuit by encoding a 2x2 image in the IBM quantum register. We used the simulator back-end for the  IBM Q 5 Tenerife. We used a simple 2x2  image as below:

\[I = 
\begin{bmatrix}
5.0 & 1.0 \\
2.0 & 3.0 \\
\end{bmatrix}
\]

We need to note the these intensities will be normalized once encoded in a quantum state. Hence, we wont be able to accurately retrieve the image. The distribution we retrieved is shown in figure 3. 

\begin{figure}
    \centering
    \begin{tikzpicture}[font=\small]
    \begin{axis}[
      ybar,
      bar width=20pt,
      xlabel={Quality},
      ylabel={Probability},
      ymin=0,
      ytick=\empty,
      xtick=data,
      axis x line=bottom,
      axis y line=left,
      enlarge x limits=0.2,
      symbolic x coords={$|00\rangle$,$|01\rangle$,$|10\rangle$,$|11\rangle$},
      xticklabel style={anchor=base,yshift=-\baselineskip},
      nodes near coords={\pgfmathprintnumber\pgfplotspointmeta\%}
    ]
      \addplot[fill=white] coordinates {
        ($|00\rangle$,40)
        ($|01\rangle$,10)
        ($|10\rangle$,30)
        ($|11\rangle$,20)
      };
    \end{axis}
 \end{tikzpicture}
    \caption{Retrieved Image as a probability distribution}
    \label{fig:retrieved}
\end{figure}
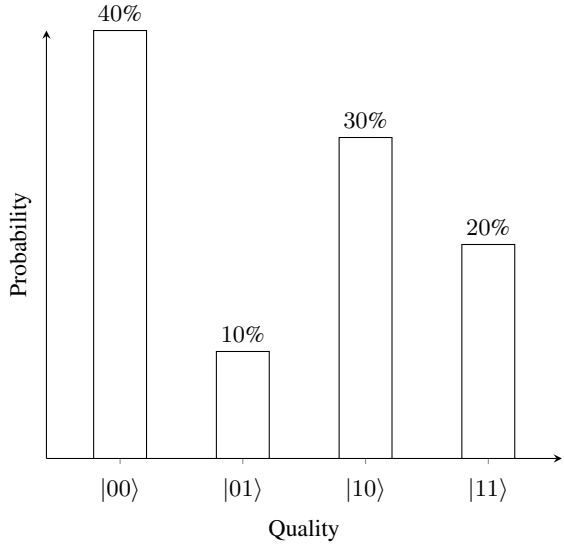

We apply usual image operations the same way as we prepare the image. Since manipulating the intensities just means controlled rotating, we can use the circuit in section 4 to target specific gates. For example, negating the image would just mean rotating the amplitude by $\pi$. The operations on a quantum image can be done by rotating the quantum bits around the Bloch sphere. For example, to invert the quantum image, we can simply rotate the qubits around the Bloch sphere by $180^o$ we get figure 3.

\begin{figure}
    \centering
    \begin{tikzpicture}[font=\small]
    \begin{axis}[
      ybar,
      bar width=20pt,
      xlabel={Quality},
      ylabel={Probability},
      ymin=0,
      ytick=\empty,
      xtick=data,
      axis x line=bottom,
      axis y line=left,
      enlarge x limits=0.2,
      symbolic x coords={$|00\rangle$,$|01\rangle$,$|10\rangle$,$|11\rangle$},
      xticklabel style={anchor=base,yshift=-\baselineskip},
      nodes near coords={\pgfmathprintnumber\pgfplotspointmeta\%}
    ]
      \addplot[fill=white] coordinates {
        ($|00\rangle$,10)
        ($|01\rangle$,40)
        ($|10\rangle$,20)
        ($|11\rangle$,30)
      };
    \end{axis}
 \end{tikzpicture}
    \caption{Retrieved Inverted Image}
    \label{fig:inverted}
\end{figure}
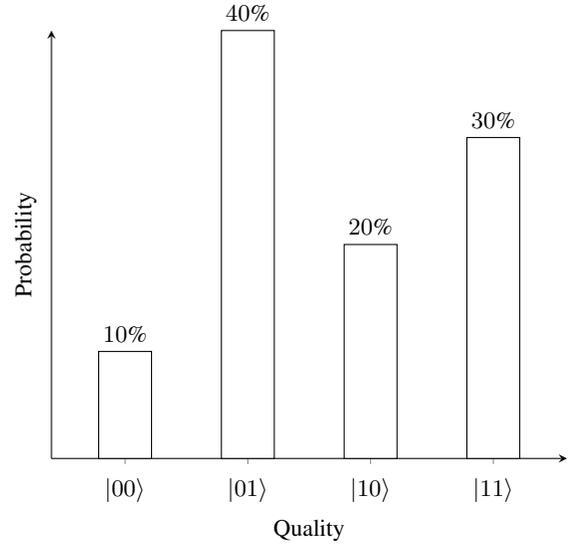

\section{Quantum Computing Frameworks}
There are several open source quantum computer simulations that function by different models of quantum computing. Table 3 lists various frameworks that the quantum algorithms can be tested on. There are also actual qubits that can be accessed via web like the IBM quantum experience. Unfortunately, we are not able to use more than 4-5 qubits, making testing of quantum models difficult. Hence, these frameworks provide simulations of quantum computers to work with. 

\begin{table}[]
    \centering
    \begin{tabular}{|c|c|}
        \hline
         Library & Environment \\
         \hline
         QETLAB  & MATLAB \\
         QISKit & Python \\
         Strawberry Fields & Python \\
         IBM Q Experience & Web GUI \\
         Quantum Dev Kit & Q\# \\
         OpenFermion & Python \\
         
         \hline
    \end{tabular}
    \caption{Some Open Source Quantum Computing Libraries}
    \label{tab:libraries}
\end{table}

\section{Conclusion and Future Work}
In this study, we explored the realm of quantum image processing and tested FRQI. We studied the implications of this field by explicitly simulating quantum behaviour on a large quantum system. We constructed a basic circuit to encode the 2x2 image. To get a more clearer picture of the mechanics of the quantum image model, we explicitly stored the quantum states classically and measured the images to demonstrate the effect of the image processing operations. In the future, we wish to expand our scope to more advanced quantum representations of images. We would also like to explore more applications like image segmentation \cite{13} and deep learning \cite{14}. The hardware side of quantum computing is being developed rapidly in this decade and there are high expectations of a commercially usable quantum computers to be a reality. This means it is important to start investigating to port classical computational problems into quantum information as soon as possible. Our future work is motivated by the promise of breaking the limitations the classical framework imposes and exploit the unique features of a quantum system discussed in section 2 to improve the efficiency of the current image processing technology. 

\section{Acknowledgements}
We thank Dr. Hugh Churchill and Arash Fereidouni from the Quantum Devices Research Lab, Department of Physics, University of Arkansas for helpful discussions and suggestions.

\end{document}